\documentclass[letterpaper, 10 pt, conference]{ieeeconf}  

\IEEEoverridecommandlockouts                              

\usepackage[top=54pt, left=54pt, right=54pt, bottom=54pt]{geometry}

\usepackage{graphicx}
\usepackage{amsmath}
\usepackage{bm}
\usepackage[algo2e]{algorithm2e} 
\usepackage{algorithm}
\usepackage{algorithmic}
\usepackage{pifont}
\usepackage{amssymb}
\usepackage{tabularx}

\usepackage{xcolor}

\DeclareMathOperator*{\argmin}{arg\,min}

\title{\LARGE \bf
Towards Scale Consistent Monocular Visual Odometry by Learning from the Virtual World
}

\author{
Sen Zhang, Jing Zhang, and Dacheng Tao
\thanks{Sen Zhang, Jing Zhang, and Dacheng Tao are with the School of Computer Science, The University of Sydney, Australia.}%
}

\begin{document}

\maketitle
\thispagestyle{empty}
\pagestyle{empty}

\begin{abstract}

Monocular visual odometry (VO) has attracted extensive research attention by providing real-time vehicle motion from cost-effective camera images. However, state-of-the-art optimization-based monocular VO methods suffer from the scale inconsistency problem for long-term predictions. Deep learning has recently been introduced to address this issue by leveraging stereo sequences or ground-truth motions in the training dataset. However, it comes at an additional cost for data collection, and such training data may not be available in all datasets. In this work, we propose VRVO, a novel framework for retrieving the absolute scale from virtual data that can be easily obtained from modern simulation environments, whereas in the real domain no stereo or ground-truth data are required in either the training or inference phases. Specifically, we first train a scale-aware disparity network using both monocular real images and stereo virtual data. The virtual-to-real domain gap is bridged by using an adversarial training strategy to map images from both domains into a shared feature space. The resulting scale-consistent disparities are then integrated with a direct VO system by constructing a virtual stereo objective that ensures the scale consistency over long trajectories. Additionally, to address the suboptimality issue caused by the separate optimization backend and the learning process, we further propose a mutual reinforcement pipeline that allows bidirectional information flow between learning and optimization, which boosts the robustness and accuracy of each other. We demonstrate the effectiveness of our framework on the KITTI and vKITTI2 datasets.

\end{abstract}

\section{INTRODUCTION}

\begin{figure}[ht!]
  \centering
  \includegraphics[width=0.45\textwidth]{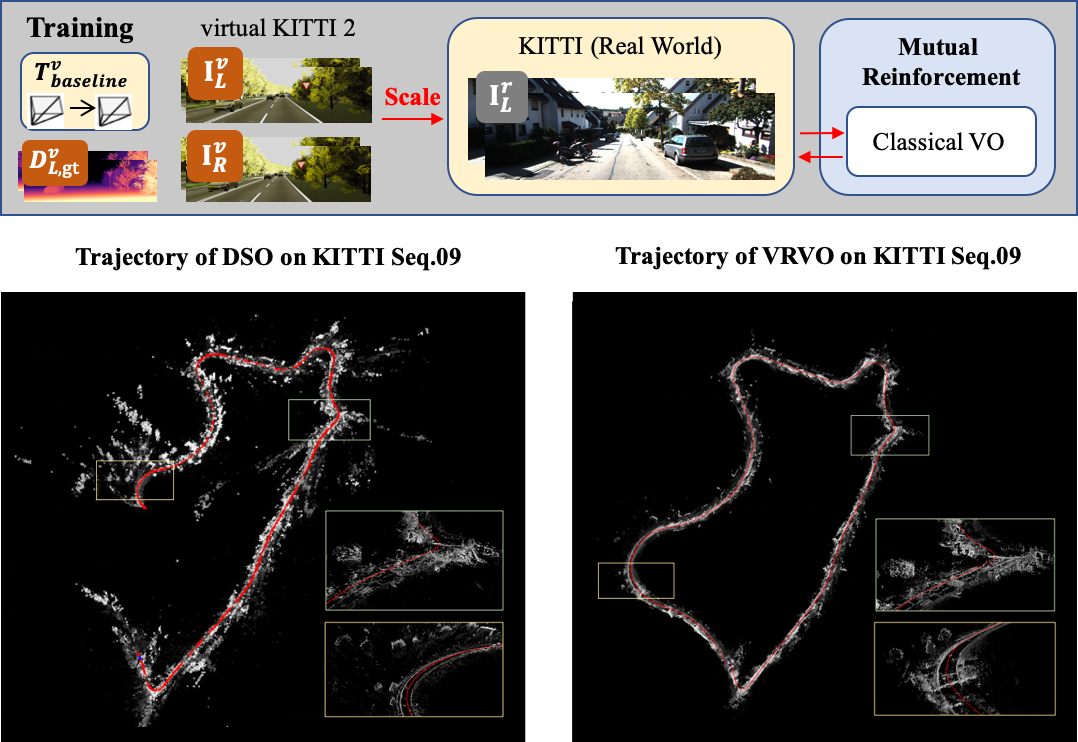}
  \caption{(a) Illustration of the proposed VRVO framework: (1) Training using virtual data provides scale information, and (2) The mutual reinforcement pipeline further improves the prediction quality using optimization feedback. (b) Example trajectories and point clouds of Seq.09 in the KITTI odometry dataset using Direct Sparse Odometry (DSO) and VRVO. It is worth noting that KITTI Seq.09 presents a closed loop trajectory. While DSO fails to close the loop and produce noisy point clouds due to the scale drift problem, VRVO significantly improves the result by leveraging the proposed domain adaptation and mutual reinforcement modules.}
  \label{illustration}
\end{figure}

Visual odometry (VO) systems play an essential role in modern robotics by providing real-time vehicle motion from visual sensors, which facilitates many downstream tasks such as autonomous driving, virtual reality, and robot manipulation~\cite{fraundorfer2012visual}\cite{zhang2020empowering}. In particular, monocular VO methods have drawn extensive research attention due to the easy setup and low cost of a single camera. The camera motion is determined by querying the geometric cues from consecutive monocular images. Previous monocular VO systems can be categorized into deep learning-based methods that directly predict camera motion by implicitly learning the geometric relationship from training data, and optimization-based methods that explicitly model the geometric equations and formulate VO as an optimization problem. While optimization-based methods achieve state-of-the-art (SOTA) performance~\cite{mur2017orb}\cite{engel2017direct}, they typically suffer from the scale inconsistency problem since the optimization objectives are equivalent up to an arbitrary scaling factor w.r.t. depth and translation, resulting in scale-inconsistent overall trajectories, as illustrated in Fig.~\ref{illustration}. 

Targeting at this issue, recent efforts have focused on integrating deep learning techniques into optimization-based pipelines to retrieve the scale information by learning from external data sources~\cite{tateno2017cnn}\cite{yang2018deep}\cite{yang2020d3vo}, or training with internal reprojected depth consistency regularization~\cite{bian2019unsupervised}\cite{zhan2020visual}. Supervised deep learning models predict camera poses and depths with absolute scale by using ground-truth labels~\cite{wang2017deepvo}\cite{xue2019beyond}. However, the performance of the learnt network is still inferior compared with optimization-based systems, partly due to the neglect of the well-established geometric relationship. Additionally, collecting ground-truth labels can also be costly and time-consuming. On the other hand, utilizing extra sensors during the training phase, such as stereo images with a known baseline, provide an alternative method for recovering the absolute scale~\cite{yang2018deep}\cite{yang2020d3vo}, while only monocular sequences are required during inference. Nevertheless, stereo images may be unavailable in real-world datasets and increase the overall cost of the data collection process.

Without ground-truth labels and stereo training data, another line of work proposes to ensure scale consistency using a local reprojected depth consistency loss~\cite{bian2019unsupervised}\cite{zhan2020visual}. Global scale consistency is then achieved by propagating the scale constraint through overlapped training clips. However, due to the propagation errors and indirect supervisory signals, these methods still perform worse than methods that utilize extra information with absolute scale.

Modern simulation engines have enabled the construction of interactive and photo-realistic virtual environments~\cite{cabon2020virtual}\cite{savva2019habitat}\cite{tartanair2020iros}, where enormous training sequences with various labels, such as depth, optical flow, surface normal, and camera motion, can be easily generated at a much lower cost, thus opening up new opportunities for resolving the inherent problems of monocular VO methods. On the other hand, current learnt networks are usually integrated into optimization-based VO systems via the predicted depth and optical flow information. However, the information flow from the learning process to the optimization backend is typically unidirectional, i.e., no feedback signals are used to supervise the learning process. This inherent separation between learning and optimization results in the suboptimality issue, which is much less explored.


To this end, we propose VRVO (Virtual-to-Real Visual Odometry), a novel and practical VO framework that requires only monocular real images in both training and inference phases, by retrieving the absolute scale from virtual data and establishing a mutual reinforcement (MR) pipeline between learning and optimization. In particular, we train a scale-aware disparity network and an auxiliary pose network using both virtual and real sequences. The virtual-to-real domain gap is bridged by mapping both virtual and real images into a shared feature space through adversarial training. Thanks to the known stereo baseline and ground-truth disparity maps in the virtual dataset, the predicted disparities are scale-aware and are fed into a direct VO system for depth initialization and the construction of an extra virtual stereo optimization objective. In contrast to previous works that have focused exclusively on the unidirectional information flow from learning to optimization, we establish the MR pipeline by using the more accurate trajectories from the optimization backend as an auxiliary regularization signal to supervise the learning process. In this way, we allow the disparity network and the VO backend to be trained and optimized in a mutually reinforced manner. We demonstrate the effectiveness of the proposed framework on virtual KITTI2~\cite{cabon2020virtual} and KITTI~\cite{geiger2013vision} w.r.t. both accuracy and robustness.


\section{Related Work}
\subsection{Scale Ambiguity of Monocular VO Systems}
Supervised pose regression methods predict absolute scale-aware motions by training the networks with ground-truth camera poses~\cite{wang2017deepvo}\cite{xue2019beyond}. However, the accuracy of pure-learning methods suffers due to the insufficient utilization of the well-established geometric constraints. Alternatively, another line of work imposes scale information on depth prediction instead, taking into account that depth and camera motion share the same scale. As such, CNN-SLAM~\cite{tateno2017cnn} integrates learnt depth maps which are trained with ground-truth depth labels into LSD-SLAM~\cite{engel2014lsd} for depth initialization. In the absence of ground-truth depth values, DVSO~\cite{yang2018deep} and D3VO~\cite{yang2020d3vo} extract the absolute scale by learning to predict both left and right disparities using stereo training sequences. The learnt disparities are then used to construct a virtual stereo optimization term for direct VO systems. Our method instead targets at the situations in which no stereo images are available in the real-world training dataset, and makes the following contributions: (1) Bridging the domain gap between virtual and real-world images to introduce the scale information learnt from the virtual world to real applications; and (2) Addressing the suboptimality issue from the separation of the optimization backend and the learning process by establishing the MR pipeline to allow bidirectional information flow between learning and optimization.

Local reprojected depth regularization provides an alternative way to ensure scale consistency using only monocular images~\cite{bian2019unsupervised}~\cite{zhao2020towards}. Nevertheless, the accuracy of these methods is still inferior to DeepVO~\cite{wang2017deepvo} and DVSO~\cite{yang2018deep}. DF-VO~\cite{zhan2020visual} incorporates this idea into an indirect VO system, utilizing a scale-consistent depth network for initialization and an optical flow network for building 2D-2D correspondences to boost the performance. In comparison, our method does not require optical flow prediction,  and thus is simpler during inference. Additionally, we address the suboptimality issue by incorporating the mutual reinforcement pipeline.

\subsection{Domain Adaptation for Depth Estimation}
Atapour et al.~\cite{atapour2018real} and T2Net~\cite{zheng2018t2net} formulated this problem as image translation from real images to the synthetic domain, and trained the depth network on synthetic datasets with ground-truth supervision. AdaDepth~\cite{kundu2018adadepth} used the adversarial approach to align the feature distributions of source and target domains and thus reduced the domain gap. The more recent GASDA \cite{zhao2019geometry} explored the setting in which stereo data are available in the real domain and added the stereo photometric loss to leverage this information. A joint synthetic-to-real and real-to-synthetic translation training scheme is proposed to enhance the results. Apart from the translation-based methods, SharinGAN~\cite{pnvr2020sharingan} mapped both virtual and real images to a shared feature space to relieve the difficulty in learning direct image translators. We follow the idea of learning a shared domain while a lightweight network structure and more informative losses are adopted~\cite{godard2019digging}. Additionally, SharinGAN also uses real stereo data during training, which is not required in our setting.

\begin{figure*}[ht!]
  \centering
  \includegraphics[width=0.85\textwidth]{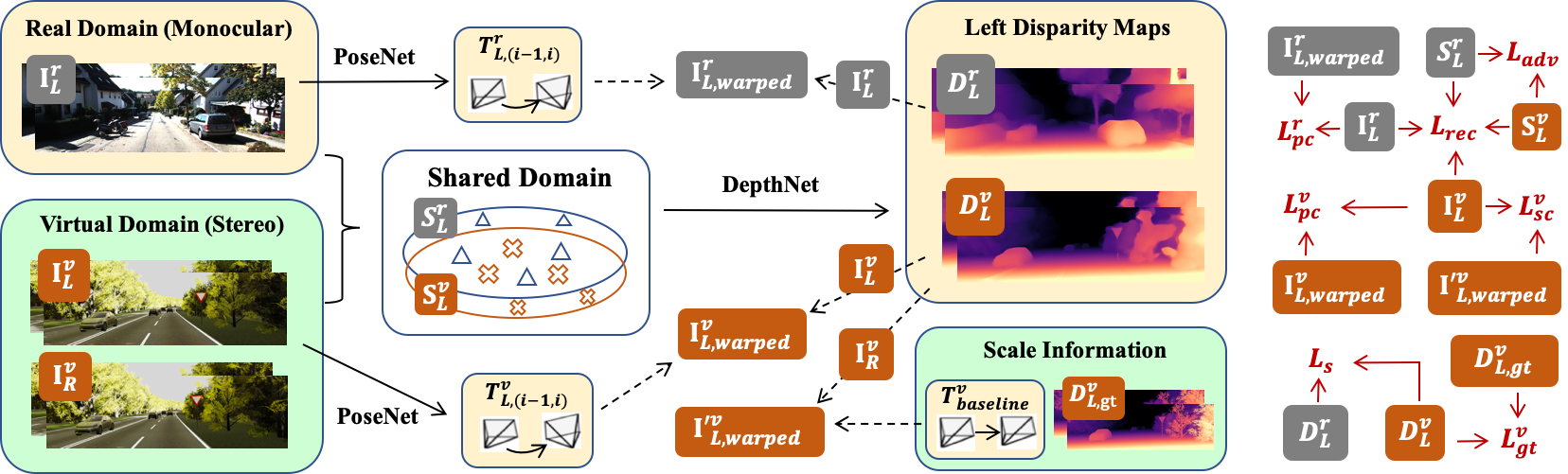}
  \caption{The overall pipeline and the losses for our domain adaptation module. The superscripts $\{r,v\}$ denote real domain and virtual domain, respectively and subscripts $\{L, R\}$ denote left image and right image, respectively. We reconstruct $I^r_{L,warped}$ and $I^v_{L,warped}$ from temporally adjacent $I^r_L$ and $I^v_L$ frames using the predicted left disparities $D^r_L$ and $D^v_L$ and the differentiable backward warping operation. We extract the absolute scale information from the virtual domain by (1) using the ground-truth depths $D^v_{L,gt}$ to provide supervision, and (2) reconstructing $I'^{v}_{L,warped}$ from the stereo image $I^v_R$ using the ground-truth stereo baseline $T^v_{baseline}$ to provide the stereo photometric consistency loss $L^v_{sc}$. }
  \label{domain_adaptation}
  \clearpage
\end{figure*}


\subsection{Supervision from Geometric VO Methods}
DVSO~\cite{yang2018deep} directly used the depth results from StereoDSO~\cite{wang2017stereo} for supervision. Klodt et al.~\cite{klodt2018supervising} used both depth and pose results from ORB-SLAM2~\cite{mur2017orb}, as well as the temporal photometric consistency loss during training. Andraghetti et al.~\cite{andraghetti2019enhancing} proposed a sparsity-invariant autoencoder to process the sparse depth maps from ORB-SLAM2~\cite{mur2017orb} and extract higher-level features. Tosi et al.~\cite{tosi2019learning} instead used the SGM stereo matching algorithm to obtain the proxy depth for supervsion. Similar to our method, Tiwari et al.~\cite{tiwari2020pseudo} proposed a self-improving loop that first performs the RGB-D version of ORB-SLAM2~\cite{mur2017orb} with depth prediction from monodepth2~\cite{godard2019digging} and then uses SLAM results as supervisory signals to finetune the depth network. Notably, our method incorporates a domain adaptation (DA) module to learn scale-aware disparities from virtual data, which are then formulated as a virtual stereo term for the optimization backend, therefore providing the absolute scale and allowing bidirectional information flow between learning and optimization.


\section{Methodology}
In this section, we present the technical details of VRVO. We first revisit the fundamentals of direct VO methods and the scale inconsistency problem. Then we turn to how VRVO solves this problem by adapting virtual scale information to real domain and addressing the suboptimality issue using the mutual reinforcement between learning and optimization.

\subsection{Direct VO Methods}
VO aims at predicting the 6-DOF relative camera pose $T=[R,t]$ from consecutive images. Direct methods formulate this problem as optimizing the photometric error between an image and its warped counterpart. Given consecutive frames $I$ and $I'$, we optimize the following objective:
\begin{equation}
    T^* = \argmin_{[R,t]} \sum_{i=1}^{N}\mathcal{L}(I'(\phi(KRK^{-1}p_i+\frac{Kt}{z_i})), I(p_i)),
    \label{eq:photometricerror}
\end{equation}
where $K$ and $N$ denote the camera intrinsics and the number of utilized pixels, $p_i$ and $z_i$ are the coordinate and corresponding depth of the selected pixel in $I$, and $R\in SO(3)$ and $t\in \mathcal{R}^3$ are the rotation matrix and the translation vector from $I$ to $I'$, respectively. $\phi(\cdot)$ and $\mathcal{L}$ denote the depth normalization and the loss function.

Equation~\eqref{eq:photometricerror} implies that $t$ and $z_i$ are actually valid up to a scaling factor. Since for VO systems we usually conduct local optimization over limited keyframes to achieve real-time performance, this scale ambiguity will result in inconsistent predictions over long trajectories, as illustrated in Fig.~\ref{illustration}(b).

\subsection{Scale-Aware Learning from Virtual Data}
Though it remains non-trivial to address the scale inconsistency problem solely from monocular training sequences, modern photorealisitic simulation engines open new opportunities by providing cost-effective training data with ground-truth labels in the virtual domain. Given that depth and translation share the same scale, we formulate scale extraction from the virtual world as the learning of a scale-aware disparity network $\mathcal{M}_D$ which is then embedded into a direct VO system~\cite{engel2017direct} to provide scale constraints.

\paragraph{Adversarial Training for Domain Adaptation}
The challenge of leveraging virtual data lies in the domain shift from virtual to real. To address this issue, we first build an end-to-end domain adaptation module that jointly learns the scale-aware disparities and narrows the domain gap for the network to work on real images. Specifically, given monocular real sequences $I^r_L$ and stereo virtual sequences $\{I^v_L,I^v_R\}$ with computer generated ground-truth baseline $t^v_{
b}$ and left disparity maps $D^v_{L,gt}$, a shared encoder $\mathcal{M}_S$ is trained to project images from both domains into a shared feature space, which is then fed into $\mathcal{M}_D$ for disparity prediction. We adopt the adversarial training strategy proposed in~\cite{pnvr2020sharingan} to align the projected features, where a discriminator $\mathcal{M}_{adv}$ is used to distinguish the projected features from two domains, by optimizing the following adversarial loss:
\begin{equation}
    \min_{\mathcal{M}_S}\max_{\mathcal{M}_{adv}} L_{adv}-\lambda_{g} L_{gp}+ L_{task} + \lambda_{r}L_{rec}, 
\end{equation}
where $L_{adv}$ is a WGAN-alike loss~\cite{arjovsky2017wasserstein} modified for shared feature encoding, $L_{gp}$ is the gradient penalty~\cite{gulrajani2017improved} to obtain more stable gradients for training $\mathcal{M}_{adv}$, $L_{rec}$ is the reconstruction loss to avoid a trivial solution of $\mathcal{M}_S$, and $L_{task}$ is the loss for scale-aware disparity prediction which will be explained in Section~III-B.b. Of note is that $\{L_{task},L_{rec}\}$ are only used for updating $\mathcal{M}_S$ in this stage.
\begin{equation}
    L_{adv}= E_{I^r_L}[\mathcal{M}_{adv}(\mathcal{M}_S(I^r_L))] - E_{I^v_L}[\mathcal{M}_{adv}(\mathcal{M}_S(I^v_L))],
\end{equation}
\begin{equation}
    L_{gp} = (||\nabla_{\Tilde{F_S}}\mathcal{M}_{adv}(\Tilde{F_S})||_2 - 1)^2, 
\end{equation}
\begin{equation}
    \Tilde{F_S} = \epsilon\mathcal{M}_S(I^r_L) + (1-\epsilon)\mathcal{M}_S(I^v_L),\epsilon\sim Uniform[0,1],
\end{equation}
\begin{equation}
    L_{rec}=||I^r_L-\mathcal{M}_S(I^r_L)||^2_2+||I^v_L-\mathcal{M}_S(I^v_L)||^2_2.
\end{equation}

\begin{figure}[t!]
  \centering
  \includegraphics[width=0.48\textwidth]{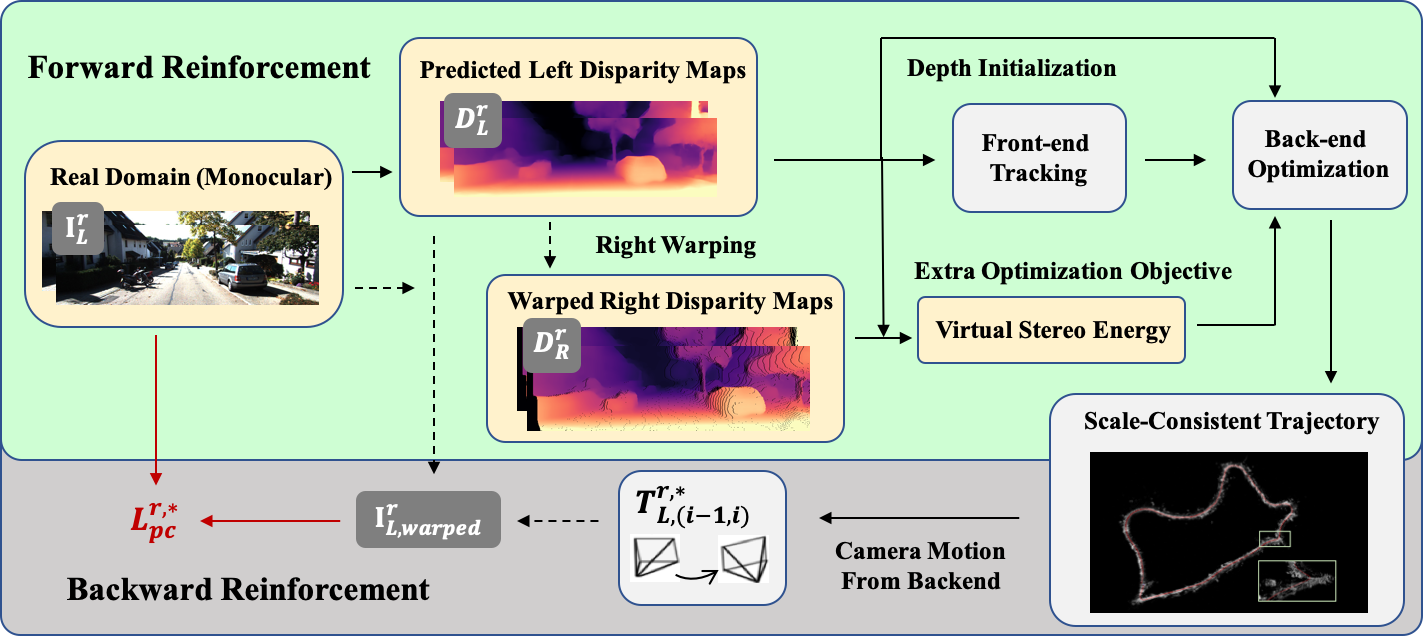}
  \caption{The pipeline of our mutual reinforcement module. The superscript $r$ denote real domain and the subscripts $\{L,R\}$ denote left and right images, respectively. $T^{r,*}_{L,(i-1,i)}$ is the relative camera motion from ($i-1$)th frame to $i$th frame in the real domain, predicted by the optimization-based VO backend. $I^r_{L,warped}$ and $L^{r,*}_{pc}$ are the reconstructed frame and the corresponding photometric consistency loss for backward reinforcement. During forward reinforcement, we generate right disparities $D^r_{R}$ from $D^r_L$ using forward warping. $D^r_R$ is then used to provide depth initialization and the virtual stereo energy term for the optimization-based backend.}
  \label{mutual_kd}
\end{figure}

\paragraph{Scale-Aware Disparity Prediction}
\label{subsec:ScaleAwareDisparityPrediction}
We train a disparity network $\mathcal{M}_D$ and an auxiliary pose network $\mathcal{M}_P$ using both unsupervised and supervised training losses from real and virtual sequences:
\begin{eqnarray}
    \min_{\mathcal{M}_D,\mathcal{M}_P} L_{task} &=&  \lambda_p(L^r_{pc} + L^v_{pc}) + \lambda_s(L^r_{s} + L^v_{s}) \nonumber \\
    & & + \lambda_{gt}L^v_{gt} + \lambda_{sc}L^v_{sc},
\end{eqnarray}
where $\{\lambda_p,\lambda_s,\lambda_{gt},\lambda_{sc}\}$ denote the weights for the corresponding loss terms. $\{L^r_{pc},L^v_{pc}\}$ and $\{L^r_{s},L^v_{s}\}$ denote the unsupervised photometric consistency losses and the disparity smoothness losses for both real and virtual images. WLOG, we omit the superscripts $r$ and $v$ for simplicity:
\begin{equation}
    L_{pc} = \frac{1}{N}\sum_{i=1}^N \min_{\delta\in\{-1,1\}} \mathcal{L}(I_L(p_i),I_{\delta}(\phi(KR_\delta K^{-1}p_i+\frac{Kt_\delta d_i}{f_xt^v_b}))),
\end{equation}
\begin{equation}
    \mathcal{L}(I_L,I_\delta) = \alpha\frac{1-SSIM(I_L,I_\delta)}{2} + (1-\alpha)||I_L-I_\delta||_1,
\end{equation}
\begin{equation}
    L_{s} = \frac{1}{N}\sum_{x,y}\sum_{a\in\{x,y\}} |\nabla_aD_L(x,y)|e^{-|\nabla_aI_L(x,y)|},
\end{equation}
where $I_\delta$ denotes the neighbor of $I_L$ with index difference $\delta$. $f_x=K[0,0]$ denotes the focal length along x axis. $[R_\delta,t_\delta]=\mathcal{M}_P([I_L,I_\delta])$ and $SSIM(\cdot)$ denote the predicted relative pose from $I_L$ to $I_\delta$ and the structural similarity index~\cite{wang2004image}, respectively. $D_L=\mathcal{M}_D(\mathcal{M}_S(I_L))$ represents the predicted disparity map where $d_i$ is the disparity of pixel $p_i$. Of note is that $t^v_b$ is the baseline in the virtual stereo setting, which is used for both $L^r_{pc}$ and $L^v_{pc}$ to ensure the scale consistency between virtual and real depth predictions.

\begin{algorithm}[t!]
\caption{The training pipeline of VRVO}
\label{domain_ada_algo}
\begin{algorithmic}[1]
    \REQUIRE Loss weights $\{\lambda^*_{p},\lambda_k | k\in\{g,r,p,s,gt,sc\}\}$; Adam optimizer $\mathcal{G}$ with hyperparamers $\{\alpha,\beta_1,\beta_2\}$.
    \REQUIRE \textbf{Real domain}: Monocular sequences $I^r_L$. 
    \REQUIRE \textbf{Virtual domain}: Stereo sequences $\{I^v_L,I^v_R\}$; 
    \REQUIRE \textbf{Virtual domain}: \text{\small Baseline $t^v_b$; Left disparities $D^v_{L,gt}$}.
    \STATE Initialize network parameters. $\{w_{n}|n\in\{adv,S,D,P\}\}$
    \FOR{$N_{tr}$ iterations}
        \STATE Sample a minibatch from $\{I^r_L,I^v_L,I^v_R,D^v_{L,gt}\}$
        \STATE Update $\mathcal{M}_{adv}$: $w_{adv} \leftarrow \mathcal{G}(\nabla_{w_{adv}} -L_{adv}+\lambda_g L_{gp})$.
        \FOR{$k_s$ steps}
            \STATE Update $\mathcal{M}_S$: $w_S \leftarrow \mathcal{G}(\nabla_{w_S} L_{adv}+L_{task}+\lambda_r L_{rec})$.
        \ENDFOR
        \STATE Update $\mathcal{M}_D,\mathcal{M}_P$: $w_D,w_P\leftarrow \mathcal{G}(\nabla_{w_D,w_P} L_{task})$.
    \ENDFOR 
    \FOR{$k_f$ steps}
        \STATE Predict left disparities $D^r_L$ of $I^r_L$ using $\mathcal{M}_D$.
        \STATE Generate right disparities $D^r_R$ by forward warping $D^r_L$.
        \STATE Generate camera motions $T^{r,*}_L$ of $I^r_L$ using an optimization-based direct VO system by 
        
        (1) Initializing depth values using $D^r_L$, and
        
        (2) Adding the virtual stereo energy $E_{vs}$.

        \FOR{$N_{ft}$ iterations}
            \STATE Sample a minibatch from $\{I^r_L,I^v_L,I^v_R,D^v_{L,gt},T^{r,*}_L\}$.
            \STATE \scalebox{0.9}{Update $\mathcal{M}_D,\mathcal{M}_P$: $w_D,w_P\leftarrow \mathcal{G}(\nabla_{w_D,w_P} L_{task}+\lambda^*_p L^{r,*}_{pc})$.}
            \ENDFOR
    \ENDFOR 
    \clearpage
\end{algorithmic}
\label{pipeline_alg}
\end{algorithm}

To empower the network with scale-aware ability, we further incorporate the supervised disparity loss $L^v_{gt}$ and the stereo consistency loss $L^v_{sc}$ into (7):
\begin{equation}
    L^v_{gt} = ||\mathcal{M}_D(\mathcal{M}_S(I^v_L)) - D^v_{L,gt}||_1,
\end{equation}
\begin{equation}
    L^v_{sc} = \frac{1}{N}\sum_{i=1}^N  \mathcal{L}(I^v_L(p_i),I^v_R(\phi(p_i+[d_i,0,0]^T))),
\end{equation}
where $D^v_{L,gt}$ denotes the ground-truth disparity map and $I^v_R$ denotes the stereo counterpart of $I^v_L$. It is worth noting that (12) implies a fixed baseline $t^v_b$ in the virtual domain, which is explicitly used in (8) for $L^r_{pc}$ to inform the depth scale.

\subsection{Mutual Reinforcement for Unified VO}
One limitation of embedding learnt depths into classical VO methods is that learning and optimization are not jointly optimized due to the indifferentiable optimization backend, resulting in a suboptimal disparity network. In this work, we unify learning and optimization by proposing a mutual reinforcement (MR) pipeline that finetunes the networks using the more accurate trajectories from the backend as supervision, thereby alleviating the suboptimality problem.

\begin{table*}[tp!]
\caption{Evaluation results on KITTI Odometry Seq.09-10. \textit{Train} denotes the training data required in the real domain, where $M$ and $S$ denote monocular and stereo sequences respectively. \textit{Online} denotes whether online parameter funetuning is required using test data. $\mathcal{M}_D$ and $\mathcal{M}_F$ denote whether the depth network and the optical flow network are required respectively. The superscript ${}^*$ and the \textbf{bold} font denote the best results among all evaluated methods, and offline methods that do not require stereo training sequences in real domain and the optical flow network $\mathcal{M}_F$, respectively.}
\label{tb:full_result}
\begin{tabular}{|c||c|c|c|c|c|c|c|c|c|c|}
\hline
 & \multicolumn{4}{c|}{} & \multicolumn{3}{c|}{Sequence 09} & \multicolumn{3}{c|}{Sequence 10}\\
\hline
Methods & \textit{Train} & \textit{Online} & $\mathcal{M}_{D}$ & $\mathcal{M}_{F}$ & $t_{err}$ ($\%$) & $r_{err}$ (${}^\circ$/100m) & ATE (m) & $t_{err}$ ($\%$) & $r_{err}$ (${}^\circ$/100m) & ATE (m)\\
\hline
 DSO~\cite{engel2017direct} & -  & -  & - & - & 15.91  & 0.20${}^*$  & 52.23  & 6.49  & 0.20${}^*$  & 11.09 \\
 \hline
 ORB-SLAM2 (w/o LC)~\cite{mur2017orb} & -  & - & - & - & 9.30  & 0.26  & 38.77  & 2.57  & 0.32  & 5.42 \\
 \hline
 ORB-SLAM2 (w/ LC)~\cite{mur2017orb} & -  & - & - & - & 2.88  & 0.25  & 8.39  & 3.30  & 0.30  & 6.63 \\
\hline
\hline
 WithoutPose~\cite{zhao2020towards} & M & & \checkmark & \checkmark & 6.93 & 0.44 & - & 4.66 & 0.62 & - \\
\hline
 DF-VO~\cite{zhan2020visual} & M &  & \checkmark & \checkmark & 2.47  & 0.30  & 11.02  & 1.96  & 0.31  & 3.37${}^*$ \\
\hline
 DF-VO~\cite{zhan2020visual} & S & & \checkmark & \checkmark & 2.61 & 0.29 & 10.88 & 2.29 & 0.37 & 3.72 \\
\hline
 DOC+~\cite{zhang2021deep} & S & \checkmark & \checkmark  &  & 2.02  & 0.61  & 4.76  & 2.29  & 1.10  & 3.38 \\
\hline
 OnlineAda-I~\cite{li2020self} & M & \checkmark & \checkmark & & 5.89 & 3.34 & - & 4.79 & 0.83 & - \\
\hline
 OnlineAda-II~\cite{li2021generalizing} & M & \checkmark & \checkmark & \checkmark & 1.87 & 0.46 & - & 1.93${}^*$ & 0.30 & - \\
\hline 
\hline
SfMLearner~\cite{zhou2017unsupervised} & M &  & \checkmark &  & 11.32  & 4.07  & 26.93  & 15.25  & 4.06  & 24.09 \\
\hline
Depth-VO-Feat~\cite{zhan2018unsupervised} & M &  & \checkmark &  & 11.89  & 3.60  & 52.12  & 12.82  & 3.41  & 24.70 \\
\hline
 SC-SfMLearner~\cite{bian2019unsupervised} & M &  & \checkmark &  & 7.64  & 2.19  & 15.02   & 10.74  & 4.58  & 20.19 \\
\hline
 DPC (w/o LC)~\cite{wagstaff2020self} & M & & \checkmark & & 2.82 & 0.76 & - & 3.81 & 1.34 & - \\
\hline 
 DPC (w/ LC)~\cite{wagstaff2020self} & M & & \checkmark & & 2.13 & 0.80 & - & 3.48 & 1.38 & - \\
\hline
 ours (w/o MR) & M &  & \checkmark &  & 1.81  & 0.30  & 5.96  & 2.78  & 0.38  & 6.26 \\
\hline 
 ours (w/ MR) & M &  & \checkmark &  & \textbf{1.55}${}^*$  & \textbf{0.28}  & \textbf{4.39}${}^*$  & \textbf{2.75}  & \textbf{0.36}  & \textbf{6.04} \\
\hline
\end{tabular}
\end{table*}

\paragraph{Forward Reinforcement}
We first use the scale-ware disparity predictions for depth initialization in a SOTA direct VO system~\cite{engel2017direct}. Following~\cite{yang2018deep}, a virtual stereo energy $E_{vs}$ is incorporated into optimization to provide scale constraints. 
\begin{equation}
    E_{vs} = \sum_{i=1}^{k} \omega_{i}||I^r_L(\phi(p^s_i-[D^r_R(p^s_i),0,0]^T))-I^r_L(p_i)||_\gamma,
\end{equation}
\begin{equation}
    p^s_i=\phi(p_iz_{i}+[f_xt^v_b,0,0]^T),
\end{equation}
where $z_i$ and $||\cdot||_\gamma$ denote the pixel depth to be optimized in the backend and the Huber norm with threshold $\gamma$, respectively. $\omega_i$ denotes the energy weight based on image gradients and $D^r_R$ denotes the disparity map of the virtual stereo counterpart, which is generated by forward warping the predicted left disparity $D^r_L=\mathcal{M}_D(\mathcal{M}_S(I^r_L))$.

\paragraph{Backward Reinforcement}
Since the optimized depth results from the backend VO system are significantly more accurate than the initial ones, we use them to provide informative supervision to further finetune the networks. Instead of regularizing $\mathcal{M}_D$ with the sparse depths from the backend, we use the optimized camera motion $T^{r,*}_L=[R^*,t^*]$ to construct a photometric regularization loss on the real domain to achieve dense supervision over $\mathcal{M}_D$ and $\mathcal{M}_P$:
\begin{equation}
    L^{r,*}_{pc} = \frac{1}{N}\sum_{i=1}^N \min_{\delta} \mathcal{L}(I^r_L(p_i),I^r_{\delta}(\phi(KR^*_\delta K^{-1}p_i+\frac{Kt^*_\delta d_i}{f_xt^v_b}))),
\end{equation}
where $\delta\in\{-1,1\}$. Since the $t^*_\delta$ from the optimization backend are already scale-consistent,
we do not use virtual data in this stage. Specifically, we fix $\{\mathcal{M}_S,\mathcal{M}_{adv}\}$ and only update $\{\mathcal{M}_D,\mathcal{M}_P\}$ using real domain sequences.

\section{Experiments}
We evaluate the effectiveness of VRVO on virtual KITTI 2 (vKITTI2)~\cite{cabon2020virtual} and KITTI odometry~\cite{geiger2013vision} autonomous driving datasets. KITTI odometry dataset contains 11 sequences collected from real-world driving scenarios with ground-truth camera motion for evaluation, and vKITTI2 provides photorealistic reconstruction of KITTI scenarios using the Unity game engine, where rich ground-truth labels such as camera pose, optical flow, and depth are available. Following the evaluation scheme in~\cite{zhan2020visual}, we test the results on sequences 09 and 10 and use the remaining monocular sequences for training, which are randomly split into 19,618 training pairs $[I^r_{L},I^r_{-1},I^r_{+1}]$ and 773 validation pairs. For vKITTI2, we use all stereo sequences for training, resulting in 20,930 training pairs $[I^v_{L},I^v_{-1},I^v_{+1},I^v_R]$. Images from both domain are cropped to 640$\times$192 during training and inference.

\subsection{Implementation Details}
We implement all networks in PyTorch~\cite{paszke2019pytorch}. $\mathcal{M}_S, \mathcal{M}_D$ and $\mathcal{M}_P$ all adopt the lightweight monodepth2~\cite{godard2019digging} network structure which uses ResNet18~\cite{he2016deep} as the backbone encoder.
We first pretrain $\{\mathcal{M}_D,\mathcal{M}_P\}$ on vKITTI2 using the raw images as  inputs, and pretrain $\mathcal{M}_S$ as a self-encoder using only $L_{rec}$. As demonstrated in Algorithm~\ref{pipeline_alg}, we then jointly train the networks by $N_{tr}=150k$ iterations with $k_s=5$, followed by $k_f=5$ MR steps. At each MR step, we run one epoch over the training sequences to update $\mathcal{M}_D$ and $\mathcal{M}_P$ while fixing $\mathcal{M}_S$ and $\mathcal{M}_{adv}$. The learning rate is set to $10^{-4}$ for domain adaptation and $10^{-3}$ for mutual reinforcement to allow jumping out of local convergence basin. $\{\lambda_g,\lambda_r\}$ are both set to 10, $\{\lambda_p,\lambda_{gt},\lambda_{sc}\}$ are all set to 1, and $\lambda_s$ is set to 0.1 throughout the training process. $\lambda^*_p$ is set to 0.01 and we also conduct ablation study on the influence of $\lambda^*_p$. The final direct VO system that uses predicted disparities for depth initialization and the virtual stereo objective is built upon the C++ implementation of DSO~\cite{engel2017direct}.

\subsection{Visual Odometry Results}
We compare VRVO with classical optimization-based methods DSO~\cite{engel2017direct} and ORB-SLAM2~\cite{mur2017orb} (with and without loop closure), end-to-end unsupervised learning methods SfMLearner~\cite{zhou2017unsupervised}, Depth-VO-Feat~\cite{zhan2018unsupervised}, SC-SfMLearner~\cite{bian2019unsupervised}, and WithoutPose~\cite{zhao2020towards}, online learning methods OnlineAda-I~\cite{li2020self}, OnlineAda-II~\cite{li2021generalizing}, and DOC+~\cite{zhang2021deep}, and the SOTA hybrid method DF-VO~\cite{zhan2020visual}. Results of two related works DVSO~\cite{yang2018deep} and D3VO~\cite{yang2020d3vo} are not presented here since KITTI odometry sequences 09 and 10 are used in their training set and both methods require stereo images in the real domain during the training phase. 

Following~\cite{zhan2020visual}, we report the average translation error $t_{err}$ ($\%$) and rotation error $r_{err}$ (${}^\circ$/100m) over all sub-sequences of lengths $\{100m, 200m, ..., 800m\}$, and the absolute trajectory error ATE (m). Due to the stochasticity of optimization, we run our direct backend five times and report the mean results. Since monocular training sequences lack absolute scale information, we apply a scale-and-align (7DOF) transformation to the results as suggested in~\cite{zhan2020visual}. 

As summarized in Table~\ref{tb:full_result}, our method achieves the best $t_{err}$ and ATE on Seq.09, and outperforms learning-based methods w.r.t. $r_{err}$ on Seq.09 as well. Though DSO achieves the best $r_{err}$ on both Seq.09 and Seq.10, it suffers from the scale inconsistency problem, resulting in unsatisfactory $t_{err}$ and ATE. Besides, the performance of our methods on Seq.10 surpasses all methods that train with monocular real sequences and do not require optical flow prediction (SfMLearner, Depth-VO-Feat, SC-SfMLearner, DPC), and is highly comparable with methods that utilize online finetuning, stereo training data, or an extra optical flow network. 

\begin{figure}[thpb]
  \centering
  \includegraphics[width=0.4\textwidth]{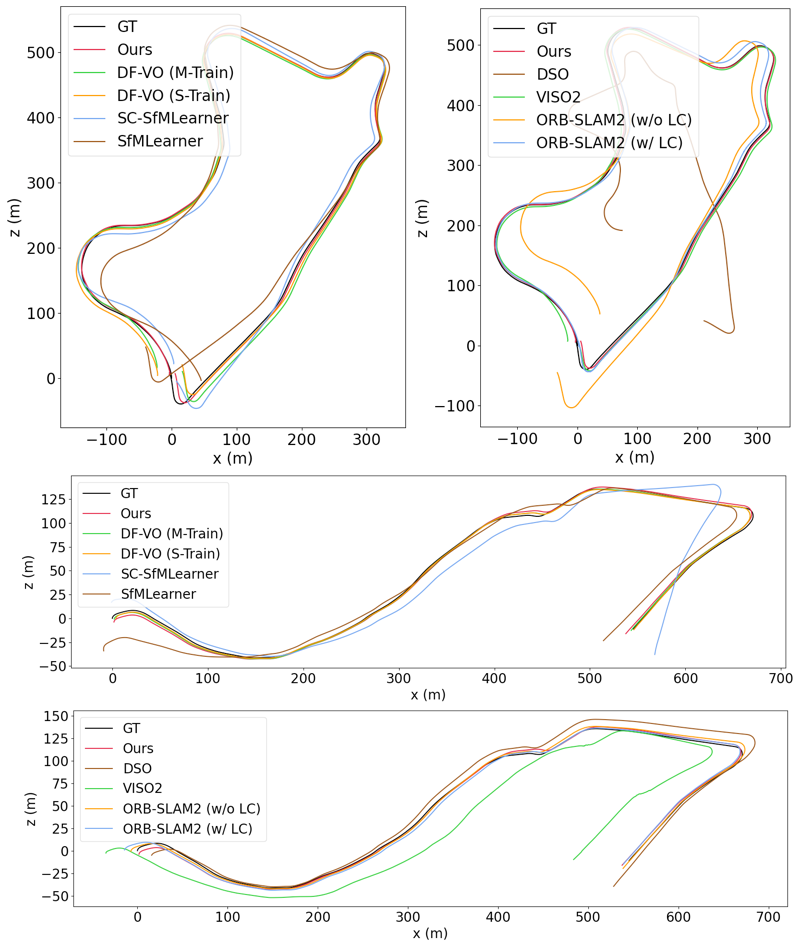}
  \caption{Predicted trajectories on KITTI odometry seq.09 (Top row) and seq.10 (Bottom rows). The results against learning-based and geometric methods are displayed separately.}
  \label{trajectory}
\end{figure}

The visualization results are presented in Fig.~\ref{trajectory}. By leveraging the scale information learnt from the virtual domain, we significantly reduce the scale inconsistency throughout the whole trajectory. Notably, we apply the same backend hyperparameter set to all sequences, which may explain the performance difference between Seq.09 and Seq.10. The design of a better and adaptive hyperparameter selection scheme presents an interesting future research direction.

\begin{table}[tp!]
\caption{Ablation studies on KITTI Odometry Seq.09. V and R denote whether the virtual and the real data are used for training, respectively. 
The \textbf{bold} metrics are reported in TABLE~\ref{tb:full_result}.}
\label{tb:abl_result}
\begin{center}
\begin{tabular}{|c|c|c|c|c|c|}
\hline
 Training & DA & MR & $\lambda^*_p$ & $t_{err}$ ($\%$) & $r_{err}$ (${}^\circ$/100m)\\
\hline
 V & & & - & 5.740$\pm$1.307 & 0.301$\pm$0.026 \\
\hline 
 R & & & - & 11.069$\pm$0.139 & 0.215$\pm$0.003 \\
\hline
 V+R &\checkmark & & - & \textbf{1.808$\pm$0.368} & \textbf{0.304$\pm$0.004} \\
\hline
 V+R &\checkmark & \checkmark & 0 & 2.024$\pm$0.121 & 0.283$\pm$0.002 \\
\hline
 V+R &\checkmark & \checkmark & 0.001 & 1.782$\pm$0.011 & 0.296$\pm$0.002 \\
\hline
 V+R &\checkmark & \checkmark & 0.01 & \textbf{1.546$\pm$0.021} & \textbf{0.280$\pm$0.002} \\
\hline
 V+R &\checkmark & \checkmark & 0.1 & 2.918$\pm$0.094 & 0.297$\pm$0.007 \\
\hline
\end{tabular}
\end{center}
\end{table}

\subsection{Ablation Studies}
We further conduct ablation studies to investigate the influence of (1) the virtual domain information, (2) the domain adaptation module, and (3) the mutual reinforcement module. Due to the space limitation, we only report results on Seq.09, as shown in Table~\ref{tb:abl_result}. We run VRVO five times at different settings and report both the means and standard errors of $t_{err}$ and $r_{err}$. As expected, $t_{err}$ of VRVO using only real domain is large due to the scale inconsistency problem. While using only virtual data with scale-aware ground-truth and baseline for training achieves a better $t_{err}$, a large standard error is observed due to the virtual-to-real domain gap. Besides, the $t_{err}$ of VRVO using only virtual domain is much worse than the domain adaptation counterpart (the third row), showing that the proposed domain adaptation module largely improves $t_{err}$ by introducing the learnt scale information into the optimization backend. Nevertheless, the standard error is still large, potentially due to the suboptimality issue caused by the inherent separation between the learning process and the optimization backend. By leveraging the proposed MR module, the standard error can be reduced significantly while the accuracy is also improved. By setting $\lambda^*_p=0$, we show that the performance gain is not achieved by further training, which instead leads to overfitting and degraded results. Besides, since the optimization results may still contain errors, a large $\lambda^*_p$ like 0.1 guides the network to overfit the intrinsic errors from the direct backend. We thus determine $\lambda^*_p$ as 0.01 for the MR stage. 

In addition, VRVO learns an absolute scale which is beyond a consistent one. The scaling ratio of the medians between the predicted depths and the ground-truth on Seq.09 and Seq.10 is 1.011 for our model with MR. We further test the generalizability of the depth network on Make3D~\cite{saxena2008make3d} and achieve a scaling ratio of 1.594, indicating that the scale-awareness can be generalized to unseen datasets. The degraded performance may come from the different camera intrinsics, which presents an interesting future research topic.


\section{Conclusion}

In this paper, we present the VRVO framework, a novel scale-consistent monocular VO system that only requires monocular real images as well as easy-to-obtain virutal data for training. It can effectively extract the scale information from the virtual data and transfer it to the real domain via a domain adaptation module and a mutual reinforcement module. Specifically, the former module learns scale-aware disparity maps while the latter one establishes bidirectional information flow between the learning process and optimization backend. Compared with SOTA monocular VO systems, our method is simpler yet achieves better results. 






\section*{ACKNOWLEDGMENT}

This work is supported by the ARC FL-170100117, IH-180100002, IC-190100031, and LE-200100049 projects.



\bibliographystyle{IEEEtran}
\bibliography{VRVO}

\end{document}